\def\year{2019}\relax
\documentclass[letterpaper]{article} 
\usepackage{aaai19}  
\usepackage{times}  
\usepackage{helvet}  
\usepackage{courier}  
\usepackage{url}  
\usepackage{graphicx}  
\usepackage{amsmath}
\usepackage{amsfonts}
\usepackage{amsthm}
\usepackage{xcolor}
\usepackage{cite}
 \usepackage{natbib}
\usepackage{graphicx,pgfplots,tikz,tikz-3dplot}
\usepackage{multirow}

\newtheorem{definition}{Definition}
\newtheorem{example}{Example}

\DeclareMathOperator*{\argmin}{arg\,min}
\DeclareMathOperator*{\argmax}{arg\,max}

\begin{document}
%
\title{A Formalization of Robustness for Deep Neural Networks}
\author{Tommaso Dreossi, Shromona Ghosh, Alberto Sangiovanni-Vincentelli, Sanjit A. Seshia\thanks{The first two authors contributed equally to this work. This work was supported in part by NSF grants 1545126 (VeHICaL), 1646208, 1739816, and 1837132, the DARPA BRASS program under agreement number FA8750-16-C0043, the DARPA Assured Autonomy program, the iCyPhy center, and Berkeley Deep Drive.}\\
University of California, Berkeley, USA
}
\maketitle
\begin{abstract}
Deep neural networks have been shown to lack robustness to small input perturbations. The process of generating the perturbations that expose the lack of robustness of neural networks is known as adversarial
input generation. This process depends on the goals and capabilities of the adversary, In this paper, we propose a unifying formalization of the adversarial input generation process from a formal methods perspective. We provide a definition of robustness that is general enough to capture different formulations. The expressiveness of our formalization is shown by modeling and comparing a variety of adversarial attack techniques.
\end{abstract}

\section{Introduction}
\label{sec:introduction}


Deep neural networks and other machine learning models have found
widespread application. However, concerns have been raised on their use in 
safety-critical systems because of their lack of robustness
to perturbations of input data.
The literature is rich in so-called ``adversarial attacks'' on neural networks where networks trained to high accuracy on data sets can be 
``fooled'' by generating inputs that differ only slightly from inputs in
the training data.
However, there is no consensus on
what constitutes robustness to adversarial attacks. 
Some papers have provided a categorization of adversarial attacks
(see~\cite{BarrenoNSJT06, PapernotMJFCS16, PapernotMSW16, GoodfellowMP18, DJS-cav18}), 
but there has been no unifying formulation  that
encompasses most, if not all, definitions of robustness in the literature.

Adversarial analysis of neural networks can be viewed as formal verification.
A typical formal verification problem consists of three components: 
the system $S$ under verification, the environment $E$ in which the system operates, and a specification $\varphi$ that formalizes the correctness 
of the system.
The problem is to check whether $S$, while operating in $E$, 
satisfies $\varphi$, usually denoted as
$S \parallel E \models \varphi$.
One of the biggest challenges for verification of 
artificial intelligence (AI) based systems, and neural networks in particular,
is to find a suitable formalization in terms of $S$, $E$, and $\varphi$~\cite{seshia-arxiv16}.

In this paper, we present such a formalization for adversarial analysis of robustness of
neural networks (and machine learning models, in general).
Specifically, 
we formulate the adversarial input generation problem as a formal verification problem where the machine learning model (deep neural network) is the system $S$ under study, the adversary (or attacker) is the environment $E$, and the robustness of the model is described by the specification $\varphi$.
Overall, we propose a unifying formulation of robustness
from a formal methods perspective that provides the following benefits:
\begin{itemize}
    \item A clear separation between the elements that form the goals of the adversary;
    \item A general formalization that facilitates comparisons among different techniques, and
    \item A bridge between adversarial analysis based on formal methods and those based on optimization and other approaches used commonly in machine learning.
\end{itemize}

In the next section, we propose our formal definition of robustness and provide examples of how our framework can be used to capture different definitions of robustness (Sec.~\ref{sec:robustness}). Next, we characterize the attackers based on strength, i.e., what level of knowledge they have about the system (Sec.~\ref{sec:attack_space}). Finally, we 
show how several existing techniques can be captured by our framework (Sec.~\ref{sec:landscape}), summarizing them in Table~\ref{tab:my_label}.

\section{Robustness as Specification}
\label{sec:robustness}

Let $\mathbb{R}$ and $\mathbb{B}$ be the sets of reals and booleans, respectively.
Let $f: X \to Y$ be a machine learning (ML) model that, for a given input $x \in X$ predicts a label $y = f(x)$. In this paper, we  
focus on deep neural networks (DNNs), albeit our approach 
extend to other ML models as well. 
Let $\tilde{X} \subseteq X$ be a set of allowed perturbed inputs,
$\mu : X \times X \to \mathbb{R}_{\geq 0}$ be a quantitative function 
(such as a distance, risk, or divergence function),
$D:(X \times X) \times \mathbb{R} \to \mathbb{B}$ be a constraint defined over $\mu$,
$A : X \times X \times \mathbb{R} \to \mathbb{B}$ be a target behavior constraint, and $\alpha,\beta \in \mathbb{R}$ be parameters. Then the problem of finding a set of inputs that falsifies the ML model can be cast as a decision problem as follows

\begin{definition}\label{def:dec_adv}
    Given $x \in X$,  
    find $x^*\in X$ such that the following constraints  hold:
    \begin{enumerate}
        \item {\em Admissibility Constraint:} $x^* \in \tilde{X}$;\label{it:adv_space}
        \item {\em Distance Constraint:} $D(\mu(x,x^*),\alpha)$, and\label{it:distance}
        \item {\em Target Behavior Constraint:} $A(x,x^*,\beta)$. \label{it:goal}
    \end{enumerate}
\end{definition}
The Admissibility Constraint~(\ref{it:adv_space}) ensures that the adversarial input $x^*$ belongs to the space of admissible perturbed inputs.  
The Distance Constraint~(\ref{it:distance}) constrains $x^*$ to be no more distant from $x$ than $\alpha$. 
Finally, the Target Behavior Constraint~(\ref{it:goal}) captures the target behavior of the adversary as
a predicate $A(x,x^*,\beta)$ which is true iff the adversary changes the behavior of the ML model by at least $\beta$ 
modifying $x$ to $x^*$.
If the three constraints hold, then we say that the ML model has failed for 
input $x$. We note that this is a so-called 
``local'' robustness property for a specific input $x$, as opposed to other notions of ``global'' robustness to changes to a population of inputs (see~\cite{DJS-cav18,SeshiaDDFGKSVY18}.

Typically, the problem of finding an adversarial example $x^*$ for a model $f$ at a given input $x \in X$ as formulated above, can be formulated as an optimization problem in one of two ways:
\begin{itemize}
    \item {\em Minimizing perturbation}: find the closest $x^*$ that alters $f$'s prediction. This can be encoded in constraint (\ref{it:distance}) as $\mu(x,x^*) \leq \alpha$;
    \item {\em Maximizing the loss}: find $x^*$ which maximizes false classification. This can be encoded in the constraint (\ref{it:goal}) as $L(f(x),f(x^*)) \geq \beta$.
\end{itemize}

\begin{definition}\label{def:opt_adv}
    The optimization version of
    Definition~\ref{def:dec_adv} is to find an input $x^*$ such that either $x^* = \argmin_{x^* \in X} \alpha$ or $x^* = \argmax_{x^* \in X} \beta$, subject to 
    the constraints in Definition~\ref{def:dec_adv}.
\end{definition}

%
The following examples demonstrate how Definition~\ref{def:dec_adv} can be used to express different formulations of the adversarial input generation.

\begin{example}
    In the seminal paper~\cite{SzegedyZSBEGF13}, the adversarial generation problem is formulated as
    $\min \| r \|_2$ subject to $f(x+r) = y$ and $x+r \in [0,1]^m$, where $r$
    is the perturbation and $y$ is the target label. This definition can be recast in our decision framework where (\ref{it:adv_space}) is $x+r \in [0,1]^m$,
    (\ref{it:distance}) is $\| r \|_2 \leq \alpha$, and
    (\ref{it:goal}) is $f(x+r) = y$.
    Here, the adversary minimizes the perturbation by solving $\argmin_{x^* \in X} \alpha$ in Definition~\ref{def:opt_adv}.
\end{example}

\begin{example}
    Madry et al.~\cite{madry2017towards} consider the robustness of models with respect to adversarial attacks. It is assumed that a loss function $L(\theta, x, y)$ is given, where $\theta \in \mathbb{R}^p$ is the set of model parameters. The paper formulates adversarial training as the robust optimization problem $\min_{\theta} \rho(\theta)$ where $\rho(\theta) = \mathbb{E}_{(x,y)\sim D}[ \max_{\delta \in S} L(\theta, x+\delta, y) ]$
    and, for each data point $x \in \mathbb{R}^d$, $S \subseteq \mathbb{R}^d$ is the set of allowed perturbations such as the $L_\infty$ ball around $x$. The inner maximization problem constitutes the adversarial attack model.
    
    In this case, the robustness problem can be encoded in
    Definition~\ref{def:dec_adv} as (\ref{it:adv_space}) $x+\delta \in \mathbb{R}^d$, (\ref{it:distance}) $\delta \in S$, and
    (\ref{it:goal}) $\max_{\delta \in S} L(\theta, x+\delta, y) \geq \beta$.
    Here, the adversary maximizes the loss by solving $\argmax_{x^* \in X} \beta$ in Definition~\ref{def:opt_adv}.
\end{example}

\begin{example}
    Yet another formulation is given by Athalye et al.~\cite{AthalyeEIK18} who address the problem of synthesizing robust adversarial examples, i.e., examples that are simultaneously adversarial over a distribution of transformations.
    The key idea is to constrain the expected effective distance between adversarial and original inputs, defined as $\mathbb{E}_{t \sim T}[d(t(x'),t(x))]$ where $T$ is a chosen distribution of transformation functions $t$ and $d$ is a distance function. Thus, for a given target label $y_t$, the problem is formulated as to find $x^* = \argmax_{x'} \mathbb{E}_{t \sim T}[\log P(y_t | t(x'))]$ subject to $\mathbb{E}_{t \sim T}[ d(t(x'),t(x))] < \epsilon$ and $x \in [0,1]^d$.
    
    In this case, the constraints of our characterization are 
    (\ref{it:adv_space}) $x^* \in [0,1]^d$,
    (\ref{it:distance}) $\mathbb{E}_{t \sim T}[ d(t(x),t(x^*)) ] \leq \epsilon$, and 
    (\ref{it:goal}) $\mathbb{E}_{t \sim T}[\log P(y_t | t(x^*))] \geq \beta$. Here, $\epsilon$ is a given constant and the adversary maximizes the loss by solving $\argmax_{x^* \in X} \beta$ in Definition~\ref{def:opt_adv}.
\end{example}

\section{Adversary as Environment}\label{sec:attack_space}


In this section, we focus on the 
environment in which the model operates, i.e., the kind of attack.


    
    
    
    
    


A key factor that determines the strength of an attack is the access that the adversary has to the model. 
There are several levels at which the attacker can operate:

\begin{itemize}
    
    \item \emph{White-box}: The adversary has access to the model's architecture. It may have full knowledge about some of the model's components such as parameters, gradients, or loss function. In many of these cases, the adversarial input generation can be recast as an optimization problem.
    
    Let $L : Y \times Y \to \mathbb{R}_{\geq 0}$ be the loss function indicating the penalty for an incorrect prediction. The adversarial attack can be formulated as
        $x^* = \argmax_{x^*} L(y,f(x^*)) \text{, subject to } \delta(z) \leq \epsilon$
    where $x^* = x + z$ and $\epsilon$ bounds the maximum perturbation applicable to $x$ in order to generate $x^*$.
    This optimization problem is typically intractable but relaxations and assumptions on $L$ and $f$ can be addressed by techniques able to efficiently generate adversarial examples. 
    
    \item \emph{Black-box}: The attacker does not have access to the model's gradients and parameters but must rely only on a limited interface that, for a given input, reveals to the adversary the model's prediction. In particular, the adversary must develop a strategy by generating a set of inputs $x_1, \dots, x_n$ and observing
    the classifications $y_1, \dots, y_n$ generated by the model. The ability of the attacker resides in observing the generated samples and identifying possible weaknesses of the model.
    \end{itemize}
    
Data poisoning or false learning~\cite{Biggio:2012} can also be seen as adversarial attacks. This family of methods falls outside the scope of this paper.

\begin{example}\label{ex:whitebox}
    \cite{Goodfellow2014ExplainingAH} assumes that the adversary has access to the gradient of the targeted model. Under the key observation that many machine learning models are linear, the proposed attack approximates the classic optimization-based adversarial attack by replacing the cost function $J(x,y)$ by a first-order Taylor series of $J(x,y)$ formed by taking the gradient at $x$. Thanks to this relaxation, the adversarial optimization problem can be solved in closed form $x^* = x + \epsilon \cdot sign(\nabla_x J(x,y))$.
\end{example}

\begin{example}\label{ex:blackbox}
    Finally, \cite{PapernotMGJCS17} assumes no knowledge about the attacked model's architecture, training data, nor training process. The only way the adversary interacts with the model is through an API that returns the predictions of the model for any input chosen by the adversary. In this technique the attacker builds a substitute model trained on a data set generated by the interaction with the original model.
    Then, the adversary, by reasoning on the substitute, develops an attack that is likely to transfer to the original model. \cite{PapernotMG16} shows how this kind of black-box attack applies across different kinds of machine learning models.
\end{example}
\section{Adversarial Landscape}
\label{sec:landscape}

In this section we survey some representative papers on adversarial input generation and cast them into our framework. Table~\ref{tab:my_label} summarizes
the various formulations surveyed here.

    


\subsection{White-Box Attacks}

Most of the proposed techniques revolve around targeted white-box attacks, 
i.e., the target behavior constraint (\ref{it:goal}) $A(x,x^*,\beta)$ is of the form $f(x^*) = y$ for a particular 
$y \neq f(x)$. These attacks are white-box in the sense that they often exploit the knowledge of the model by 
using gradient based optimization. 

The seminal work~\cite{SzegedyZSBEGF13},  showed that, by adapting the L-BFGS method to a box-constrained optimization problem, imperceptible perturbations are sufficient to make neural networks fail. In~\cite{PapernotMJFCS16}, the authors formalize the space of adversaries against NN and introduce a novel class of algorithms to craft adversarial samples based on a precise understanding of the mapping between inputs and outputs of NNs, referred to as Jacobian Saliency Map Attacks (JSMA). Recent improvements~\cite{Carlini017} exploiting the Adam optimizer further reduced the perturbations finding attacks that can break defensive distillations. Recently~\cite{AthalyeC018} proposed a generalized attack against networks with obfuscated gradients, where the existing gradient based attacks techniques performs poorly. This new attack technique, Backward Pass Differentiable Approximation (BPDA) builds a differentiable approximation of the layers of the NN to find adversarial examples. 

Some approaches address more formally the adversarial generation problem by specializing over particular classes of models.
For instance, \cite{WongK18} addresses ReLU neural networks and proposes a linear program based optimization procedure that minimizes the worst case loss over a convex approximation of the set of activations reachable through bounded perturbations.
A different formulation given by~\cite{ChenSZYH18}, addresses an elastic-net optimization formulation which involves an objective function containing a regularizer that linearly combines $L^1$ and $L^2$ penalty functions.


An orthogonal problem to finding the adversarial examples is verifying that no adversarial examples exist in a neighbourhood of an input. A common way to characterize this neighbourhood is by defining a ball around the input, whose radius is defined as the robustness metric.
In these cases, the distance constraint (\ref{it:distance}) $D(\mu(x,x^*), \alpha)$ takes the form $\| x-x^* \| \leq \alpha$,
where $\|\cdot\|$ is a norm (often $L^0$ or $L^\infty$) and the target behaviour $A(x,x^*,\beta)$ becomes
$\forall x^* (f(x^*) = f(x))$.

In~\cite{weng2018towards}, the authors propose a linear and Lipschitz approximation of ReLU networks. These approximations can be used to compute a tight lower bound of the robustness by efficiently propagating the bounds through the NN layers using matrix products.
The authors in~\cite{dvijotham2018dual} formulate verification is an optimization problem, and solve a Lagrangian relaxation of the optimization problem to obtain an upper bound on the worst case violation within a prespecified neighborhood around an input. 



We now look at non-targeted white-box attacks where the adversary's target behaviour $A(x,x^*,\beta)$ is relaxed to $f(x^*) \neq f(x)$.
Non-targeted adversarial examples are easier to find as compared to their targeted counterpart, but harder to defend against.

We already encountered a non-targeted white-box attack in Ex.~\ref{ex:whitebox} where the authors~\cite{Goodfellow2014ExplainingAH} exploit the linear nature of NN to identify adversarial examples.
DeepFool~\cite{Moosavi-Dezfooli16} is another approach based on an iterative linearization of the classifier. It finds adversarial examples by generating minimal perturbations that are sufficient to change classification labels. In~\cite{madry2017towards} the authors provide a saddle point formulation for training a NN against adversarial examples. They conjecture that the adversarial examples found by projected gradient descent (PGD) are the strongest first-order adversaries. They showed that in order to obtain a model robust against all first-order adversarial examples,
it is sufficient to increase the capacity of the network and train it on the adversarial examples found by PGD.

\cite{weng2018evaluating} convert the robustness analysis of a NN into a local Lipschitz constant estimation problem, and propose to use the Extreme Value Theory for efficient evaluation of the Lipschtiz constant. Their analysis yields a novel attack-agnostic robustness metric, Cross Lipschitz Extreme Value for nEtwork Robustness~(CLEVER), which is computed using gradients from i.i.d. samples with backpropagation.
\cite{ZantedeschiNR17} propose building more robust NN by introducing two new constraints, namely restricting the activation functions to bounded ReLU, and training on Gaussian augmented data, i.e., training data augmented with inputs perturbed with Gaussian noise. The overall effect is a smoother, more stable model, which is able to sustain a wide range of adversarial attacks (both white-box and black-box). This technique avoids computing adversarial examples and training on them.

Similarly to the targeted attacks case, 
a range of verification techniques have been proposed for the class of non-targeted attacks,
where the adversary's target behaviour $A(x,x^*,\beta)$ is of the form $\forall x^* (f(x^*) = y$).

Reluplex~\cite{KatzBDJK17} extends the simplex method to handle non-convex ReLU activation functions to verify local robustness of a NN. In~\cite{HuangKWW17}, the authors propose a general framework for verifying NN without restricting the activation functions. They discretize the neighborhood around a given input, and exhaustively search it with an SMT solver for an adversarial misclassification. The authors in~\cite{DuttaJST18} employ a local gradient search and a global mixed-integer optimization program to compute the output range for ReLU networks for polytopic neighbourhoods of the input. In~\cite{RuanHK18} robustness analysis is recast into a reachability problem which is solved using an adaptive nested optimization technique. Finally, in~\cite{Ehlers17} generate a linear approximation of NN with piece-wise linear activation functions.


\subsection{Black-Box Attacks}

We now look at a few black-box attacks where the adversary does not have access to the model's data, training process, nor architecture.

In Ex.~\ref{ex:blackbox} we mentioned how~\cite{PapernotMG16} exposes the strong phenomenon of adversarial sample transferability across models trained by different techniques.
In~\cite{PapernotMGJCS17}, the authors generate adversarial examples from a substitute model $f_{sub}$ trained on synthetic data. This approach, called Jacobian-based Dataset Augmentation, generates inputs using the Jacobian of the NN. This allows us to build a substitute NN accurately approximating the original network's decision boundaries using far lesser samples.
If the substitute model $f_{sub}$ accurately captures the original model $f$, then
an adversarial example $x^*$ generated for $f_{sub}$ should be transferable
to $f$, i.e., $f_{sub}(x^*) \neq f_{sub}(x) \implies f(x^*) \neq f(x)$.

Another ensemble based approach is shown in~\cite{LiuCLS16}, where the authors train multiple NN using the data collected from querying the target network. An alternate approach, which does not depend on training a substitute NN is introduced in~\cite{ChenZSYH17}. The authors propose a zero-th order optimization (ZOO) based attacks to directly estimate the gradients of the targeted NN for generating adversarial examples. Finally, genetic programming and black-box morphing agents have been considered for finding adversarial examples in \cite{XuQE16} and \cite{DangHC17} respectively.

A special type of black-box attack, model-extraction attacks, is shown in~\cite{TramerZJRR16} where the adversary with black-box access, but no prior knowledge of an ML model's parameters or training data, aims to duplicate the functionality of (i.e., "steal") the model.~\cite{Xu0Q18} propose feature squeezing as a technique to harden NN models by detecting adversarial examples. Feature squeezing reduces the search space available to an adversary by coalescing samples that correspond to many different feature vectors in the original space into a single sample.

\subsection{Semantic Perturbations}

In the works that have been presented so far, we do not consider the context of the perturbation or the adversarial example. This renders the adversarial sample to be somewhat ad hoc and may not capture realistic samples. We now explore some works which consider semantics in the adversarial example generation process.
In our general robustness framework, this corresponds to forcing the adversary to sample from a particular set of
allowed perturbations $\tilde{X} \subseteq X$ or dealing with special distance constraints $D(\mu(x,x^*), \alpha)$ that quantify the semantic similarity of two inputs rather than just their pure representation.

DeepXplore~\cite{PeiCYJ17}, is a white-box framework for systematically testing real-world deep learning systems. The interesting aspect of DeepXplore is that it produces adversarial examples by altering a particular input $x$ with meaningful perturbations $T(x)$ (e.g., brightness adjusting or occlusion) rather than just noise. In addition, DeepXplore compares examples using a coverage metric called neuron coverage $f_n(x)$ that quantifies how many neurons are activated by an input. DeepXplore searches for inputs that achieve high neuron coverage, i.e., $\max_{x^* \in T(x)} f_n(x^*)$. 

\cite{AthalyeEIK18} looks at generating targeted robust adversarial examples for NN. The authors generate adversarial examples which remain adversarial over a distribution of transformations such as translation, rotation and 3D-rendering.
Another notable work in this space, which considers verification is~\cite{WickerHK18}, use SIFT (Scale Invariant Feature Transform) to extract features from the input image. They formalize adversarial example generation as a stochastic two player game; where player 1 choose the feature to modify, and player 2 chooses the associated pixel and perturbation. 

While these techniques studied how semantics affects the search or test metric, recent work has also looked at defining robustness in terms of an abstract semantic feature space.
The authors of~\cite{DreossiDS17, dreossi-rmlw17, DJS-cav18} assume that 
the attacker is equipped with a renderer $R : S \to X$ that maps an abstract representation of the world (e.g., the pose of objects composing a scene) into a concrete input for the analyzed model (e.g., a picture of the scene) and a closed loop system model $M : S \to \mathbb{R}^n$. Thus, the search of adversarial examples is performed on the abstract space $S$ that is equipped with a function $\mu : S \times S \to \mathbb{R}$ that quantifies the semantic similarity between abstract items.
An input is considered to be adversarial if it leads a system $M_f$ using an ML model $f$ to violate a system-level specification $\phi$. This means that the adversary does not focus on the ML model only, but rather on the falsification of the specification at the system level. The target of the adversary is to find a diverse set of adversarial examples $\{x_1^*, x_2^*, \dots \}$ such that $\forall x^*_i, M(x_i^*) \not\models \varphi$. Diversity can be captured in the distance constraint as 
$\mu(R^{-1}(x_i^*),R^{-1}(x_j^*),) \leq \alpha$. In this particular case, the adversary is maximizing the $\alpha$ parameter in Definition~\ref{def:opt_adv}.
Such an approach can be useful to improve the accuracy of a neural network by augmenting training sets~\cite{DreossiGYKSS18}.

\begin{table*}[]
    \centering
    \resizebox{\textwidth}{!}{
    \begin{tabular}{c|c|c|c}
        \bf Paper & $\mathbf{x^* \in X}$ & $\mathbf{D(\mu(x,x^*), \alpha)}$ & $\mathbf{A(x,x^*, \beta)}$  \\
        \hline
        \cite{SzegedyZSBEGF13}  & \multirow{4}{*}{$x^* = x+r \in X$}             & \multirow{4}{*}{$\|r\|_p \leq \alpha $}& \multirow{4}{*}{$f(x^*)=y$} \\
        \cite{Goodfellow2014ExplainingAH} & & &\\
        \cite{papernot2016distillation} & & &\\
        \cite{Carlini017} & & &\\
        \hline
        \cite{Moosavi-Dezfooli16} & \multirow{3}{*}{$x^* = x+r \in X$} & \multirow{3}{*}{$\|r\|_p \leq \alpha$} & \multirow{3}{*}{$f(x^*) \neq f(x)$} \\
        \cite{AthalyeC018} & & & \\
        \cite{weng2018towards} & & & \\
        \hline
        \cite{madry2017towards} & $x^* = x+r \in X$   & $\| r \|_\infty \leq \alpha = \epsilon$ & $L(\theta, x^*, y) \geq \beta$\\
        \hline
        \cite{AthalyeEIK18} & $x^* = x+r \in X$ & $\mathbb{E}_{t \sim T}[ d(t(x^*),t(x)) ] \leq \alpha = \epsilon$ & $\mathbb{E}_{t \sim T}[\log P(y_t | t(x^*))] \geq \beta$\\
        \hline
        \cite{dvijotham2018dual} & \multirow{5}{*}{$x^* = x + r \in X$} & \multirow{5}{*}{$x^* \in S_{in}(x)$} & \multirow{5}{*}{$f(x^{*}) \not\in S_{out}(f(x))$}\\
        \cite{KatzBDJK17} & & & \\
        \cite{HuangKWW17} & & & \\
        \cite{DuttaJST18} & & & \\
        \cite{RuanHK18} & & & \\
        \hline 
        \cite{LiuCLS16} & \multirow{3}{*}{$x^* = x + r \in X$} & \multirow{3}{*}{$\| r \|_2 \leq \alpha$} & \multirow{3}{*}{\shortstack{$f_{sub}(x^*) = y, f_{sub}(x^*) \neq f_{sub}(x)$\\$f_{sub}(x) \neq f_{sub}(x^*) \implies f(x) \neq f(x^*)$}}\\
        \cite{PapernotMGJCS17} & & & \\
        \cite{TramerZJRR16} & & & \\
        \hline 
        \multirow{2}{*}{\cite{PeiCYJ17}} & \multirow{2}{*}{$x^* \in X$} & \multirow{2}{*}{$x^* \in \{\gamma x, x+r \} $} & $f_1(x) = \dots = f_k(x) \implies f_i(x^*) \neq f_j(x^*)$\\
         & & & $F_n(x^*) \geq \beta$\\
        \hline
        \cite{DreossiDS17} & $s^* \in S \implies x^* = R(s^*) \in X$ & $\mu(s^*_i, s^*_j) \leq \alpha$ & $f(R(s^*)) \implies M(s^*) \not\models \varphi$\\
    \end{tabular}}
    \caption{Different adversary input generation techniques under the same general notion of robustness.}
    \label{tab:my_label}
\end{table*}

\section{Conclusion\label{sec:conclusion}}

In this paper we proposed a general formal definition of robustness of neural networks. We showed how our framework can be used to capture different adversarial input generation techniques. Our work is part of a broader effort to formalize properties of neural networks (see~\cite{SeshiaDDFGKSVY18}) such as input-output relations, semantic invariance, and fairness.

\bibliography{biblio.bib}

\begin{thebibliography}{}

\bibitem[\protect\citeauthoryear{Athalye \bgroup \em et al.\egroup
  }{2018a}]{AthalyeC018}
Anish Athalye, Nicholas Carlini, and David~A. Wagner.
\newblock Obfuscated gradients give a false sense of security: Circumventing
  defenses to adversarial examples.
\newblock In {\em Proceedings of the 35th International Conference on Machine
  Learning, {ICML}}, 2018.

\bibitem[\protect\citeauthoryear{Athalye \bgroup \em et al.\egroup
  }{2018b}]{AthalyeEIK18}
Anish Athalye, Logan Engstrom, Andrew Ilyas, and Kevin Kwok.
\newblock Synthesizing robust adversarial examples.
\newblock In {\em Proceedings of the 35th International Conference on Machine
  Learning, {ICML}}, 2018.

\bibitem[\protect\citeauthoryear{Barreno \bgroup \em et al.\egroup
  }{2006}]{BarrenoNSJT06}
Marco Barreno, Blaine Nelson, Russell Sears, Anthony~D. Joseph, and J.~D.
  Tygar.
\newblock Can machine learning be secure?
\newblock In {\em Proceedings of the 2006 {ACM} Symposium on Information,
  Computer and Communications Security, {ASIACCS}}, 2006.

\bibitem[\protect\citeauthoryear{Biggio \bgroup \em et al.\egroup
  }{2012}]{Biggio:2012}
Battista Biggio, Blaine Nelson, and Pavel Laskov.
\newblock Poisoning attacks against support vector machines.
\newblock In {\em Proceedings of the 29th International Conference on
  International Conference on Machine Learning}, ICML'12, pages 1467--1474,
  USA, 2012. Omnipress.

\bibitem[\protect\citeauthoryear{Carlini and Wagner}{2017}]{Carlini017}
Nicholas Carlini and David~A. Wagner.
\newblock Towards evaluating the robustness of neural networks.
\newblock In {\em 2017 {IEEE} Symposium on Security and Privacy, {SP}}, 2017.

\bibitem[\protect\citeauthoryear{Chen \bgroup \em et al.\egroup
  }{2017}]{ChenZSYH17}
Pin{-}Yu Chen, Huan Zhang, Yash Sharma, Jinfeng Yi, and Cho{-}Jui Hsieh.
\newblock {ZOO:} zeroth order optimization based black-box attacks to deep
  neural networks without training substitute models.
\newblock In {\em Proceedings of the 10th {ACM} Workshop on Artificial
  Intelligence and Security, AISec@CCS}, 2017.

\bibitem[\protect\citeauthoryear{Chen \bgroup \em et al.\egroup
  }{2018}]{ChenSZYH18}
Pin{-}Yu Chen, Yash Sharma, Huan Zhang, Jinfeng Yi, and Cho{-}Jui Hsieh.
\newblock {EAD:} elastic-net attacks to deep neural networks via adversarial
  examples.
\newblock In {\em Proceedings of the Thirty-Second {AAAI} Conference on
  Artificial Intelligence}, 2018.

\bibitem[\protect\citeauthoryear{Dang \bgroup \em et al.\egroup
  }{2017}]{DangHC17}
Hung Dang, Yue Huang, and Ee{-}Chien Chang.
\newblock Evading classifiers by morphing in the dark.
\newblock In {\em Proceedings of the 2017 {ACM} {SIGSAC} Conference on Computer
  and Communications Security, {CCS}}, 2017.

\bibitem[\protect\citeauthoryear{Dreossi \bgroup \em et al.\egroup
  }{2017a}]{DreossiDS17}
Tommaso Dreossi, Alexandre Donz{\'{e}}, and Sanjit~A. Seshia.
\newblock Compositional falsification of cyber-physical systems with machine
  learning components.
\newblock In {\em {NASA} Formal Methods - 9th International Symposium, {NFM}},
  2017.

\bibitem[\protect\citeauthoryear{Dreossi \bgroup \em et al.\egroup
  }{2017b}]{dreossi-rmlw17}
Tommaso Dreossi, Shromona Ghosh, Alberto~L. Sangiovanni{-}Vincentelli, and
  Sanjit~A. Seshia.
\newblock Systematic testing of convolutional neural networks for autonomous
  driving.
\newblock In {\em ICML Workshop on Reliable Machine Learning in the Wild
  (RMLW)}, 2017.

\bibitem[\protect\citeauthoryear{Dreossi \bgroup \em et al.\egroup
  }{2018a}]{DreossiGYKSS18}
Tommaso Dreossi, Shromona Ghosh, Xiangyu Yue, Kurt Keutzer, Alberto~L.
  Sangiovanni{-}Vincentelli, and Sanjit~A. Seshia.
\newblock Counterexample-guided data augmentation.
\newblock In {\em Proceedings of the Twenty-Seventh International Joint
  Conference on Artificial Intelligence, {IJCAI}}, 2018.

\bibitem[\protect\citeauthoryear{Dreossi \bgroup \em et al.\egroup
  }{2018b}]{DJS-cav18}
Tommaso Dreossi, Somesh Jha, and Sanjit~A. Seshia.
\newblock Semantic adversarial deep learning.
\newblock In {\em 30th International Conference on Computer Aided Verification
  (CAV)}, 2018.

\bibitem[\protect\citeauthoryear{Dutta \bgroup \em et al.\egroup
  }{2018}]{DuttaJST18}
Souradeep Dutta, Susmit Jha, Sriram Sankaranarayanan, and Ashish Tiwari.
\newblock Output range analysis for deep feedforward neural networks.
\newblock In {\em {NASA} Formal Methods - 10th International Symposium, {NFM}},
  2018.

\bibitem[\protect\citeauthoryear{Dvijotham \bgroup \em et al.\egroup
  }{2018}]{dvijotham2018dual}
Krishnamurthy Dvijotham, Robert Stanforth, Sven Gowal, Timothy~A. Mann, and
  Pushmeet Kohli.
\newblock A dual approach to scalable verification of deep networks.
\newblock {\em CoRR}, abs/1803.06567, 2018.

\bibitem[\protect\citeauthoryear{Ehlers}{2017}]{Ehlers17}
R{\"{u}}diger Ehlers.
\newblock Formal verification of piece-wise linear feed-forward neural
  networks.
\newblock In {\em Automated Technology for Verification and Analysis - 15th
  International Symposium, {ATVA}}, 2017.

\bibitem[\protect\citeauthoryear{Goodfellow \bgroup \em et al.\egroup
  }{2014}]{Goodfellow2014ExplainingAH}
Ian~J. Goodfellow, Jonathon Shlens, and Christian Szegedy.
\newblock Explaining and harnessing adversarial examples.
\newblock {\em CoRR}, abs/1412.6572, 2014.

\bibitem[\protect\citeauthoryear{Goodfellow \bgroup \em et al.\egroup
  }{2018}]{GoodfellowMP18}
Ian~J. Goodfellow, Patrick~D. McDaniel, and Nicolas Papernot.
\newblock Making machine learning robust against adversarial inputs.
\newblock 2018.

\bibitem[\protect\citeauthoryear{Huang \bgroup \em et al.\egroup
  }{2017}]{HuangKWW17}
Xiaowei Huang, Marta Kwiatkowska, Sen Wang, and Min Wu.
\newblock Safety verification of deep neural networks.
\newblock In {\em Computer Aided Verification - 29th International Conference,
  {CAV}}, 2017.

\bibitem[\protect\citeauthoryear{Katz \bgroup \em et al.\egroup
  }{2017}]{KatzBDJK17}
Guy Katz, Clark~W. Barrett, David~L. Dill, Kyle Julian, and Mykel~J.
  Kochenderfer.
\newblock Reluplex: An efficient {SMT} solver for verifying deep neural
  networks.
\newblock In {\em Computer Aided Verification - 29th International Conference,
  {CAV}}, 2017.

\bibitem[\protect\citeauthoryear{Liu \bgroup \em et al.\egroup
  }{2016}]{LiuCLS16}
Yanpei Liu, Xinyun Chen, Chang Liu, and Dawn Song.
\newblock Delving into transferable adversarial examples and black-box attacks.
\newblock {\em CoRR}, abs/1611.02770, 2016.

\bibitem[\protect\citeauthoryear{Madry \bgroup \em et al.\egroup
  }{2017}]{madry2017towards}
Aleksander Madry, Aleksandar Makelov, Ludwig Schmidt, Dimitris Tsipras, and
  Adrian Vladu.
\newblock Towards deep learning models resistant to adversarial attacks.
\newblock {\em CoRR}, abs/1706.06083, 2017.

\bibitem[\protect\citeauthoryear{Moosavi{-}Dezfooli \bgroup \em et al.\egroup
  }{2016}]{Moosavi-Dezfooli16}
Seyed{-}Mohsen Moosavi{-}Dezfooli, Alhussein Fawzi, and Pascal Frossard.
\newblock Deepfool: {A} simple and accurate method to fool deep neural
  networks.
\newblock In {\em 2016 {IEEE} Conference on Computer Vision and Pattern
  Recognition, {CVPR}}, 2016.

\bibitem[\protect\citeauthoryear{Papernot \bgroup \em et al.\egroup
  }{2016a}]{PapernotMG16}
Nicolas Papernot, Patrick~D. McDaniel, and Ian~J. Goodfellow.
\newblock Transferability in machine learning: from phenomena to black-box
  attacks using adversarial samples.
\newblock {\em CoRR}, abs/1605.07277, 2016.

\bibitem[\protect\citeauthoryear{Papernot \bgroup \em et al.\egroup
  }{2016b}]{PapernotMJFCS16}
Nicolas Papernot, Patrick~D. McDaniel, Somesh Jha, Matt Fredrikson, Z.~Berkay
  Celik, and Ananthram Swami.
\newblock The limitations of deep learning in adversarial settings.
\newblock In {\em {IEEE} European Symposium on Security and Privacy,
  EuroS{\&}P}, 2016.

\bibitem[\protect\citeauthoryear{Papernot \bgroup \em et al.\egroup
  }{2016c}]{PapernotMSW16}
Nicolas Papernot, Patrick~D. McDaniel, Arunesh Sinha, and Michael~P. Wellman.
\newblock Towards the science of security and privacy in machine learning.
\newblock {\em CoRR}, abs/1611.03814, 2016.

\bibitem[\protect\citeauthoryear{Papernot \bgroup \em et al.\egroup
  }{2016d}]{papernot2016distillation}
Nicolas Papernot, Patrick~D. McDaniel, Xi~Wu, Somesh Jha, and Ananthram Swami.
\newblock Distillation as a defense to adversarial perturbations against deep
  neural networks.
\newblock In {\em {IEEE} Symposium on Security and Privacy, {SP}}, 2016.

\bibitem[\protect\citeauthoryear{Papernot \bgroup \em et al.\egroup
  }{2017}]{PapernotMGJCS17}
Nicolas Papernot, Patrick~D. McDaniel, Ian~J. Goodfellow, Somesh Jha, Z.~Berkay
  Celik, and Ananthram Swami.
\newblock Practical black-box attacks against machine learning.
\newblock In {\em Proceedings of the 2017 {ACM} on Asia Conference on Computer
  and Communications Security, AsiaCCS}, 2017.

\bibitem[\protect\citeauthoryear{Pei \bgroup \em et al.\egroup
  }{2017}]{PeiCYJ17}
Kexin Pei, Yinzhi Cao, Junfeng Yang, and Suman Jana.
\newblock Deepxplore: Automated whitebox testing of deep learning systems.
\newblock In {\em Proceedings of the 26th Symposium on Operating Systems
  Principles}, 2017.

\bibitem[\protect\citeauthoryear{Ruan \bgroup \em et al.\egroup
  }{2018}]{RuanHK18}
Wenjie Ruan, Xiaowei Huang, and Marta Kwiatkowska.
\newblock Reachability analysis of deep neural networks with provable
  guarantees.
\newblock In {\em Proceedings of the Twenty-Seventh International Joint
  Conference on Artificial Intelligence, {IJCAI}}, 2018.

\bibitem[\protect\citeauthoryear{Seshia \bgroup \em et al.\egroup
  }{2016}]{seshia-arxiv16}
Sanjit~A. Seshia, Dorsa Sadigh, and S.~Shankar Sastry.
\newblock {Towards Verified Artificial Intelligence}.
\newblock {\em ArXiv e-prints}, July 2016.

\bibitem[\protect\citeauthoryear{Seshia \bgroup \em et al.\egroup
  }{2018}]{SeshiaDDFGKSVY18}
Sanjit~A. Seshia, Ankush Desai, Tommaso Dreossi, Daniel~J. Fremont, Shromona
  Ghosh, Edward Kim, Sumukh Shivakumar, Marcell Vazquez{-}Chanlatte, and
  Xiangyu Yue.
\newblock Formal specification for deep neural networks.
\newblock In {\em 16th International Symposium on Automated Technology for
  Verification and Analysis (ATVA)}, pages 20--34, 2018.

\bibitem[\protect\citeauthoryear{Szegedy \bgroup \em et al.\egroup
  }{2013}]{SzegedyZSBEGF13}
Christian Szegedy, Wojciech Zaremba, Ilya Sutskever, Joan Bruna, Dumitru Erhan,
  Ian~J. Goodfellow, and Rob Fergus.
\newblock Intriguing properties of neural networks.
\newblock {\em CoRR}, abs/1312.6199, 2013.

\bibitem[\protect\citeauthoryear{Tram{\`{e}}r \bgroup \em et al.\egroup
  }{2016}]{TramerZJRR16}
Florian Tram{\`{e}}r, Fan Zhang, Ari Juels, Michael~K. Reiter, and Thomas
  Ristenpart.
\newblock Stealing machine learning models via prediction apis.
\newblock In {\em 25th {USENIX} Security Symposium, {USENIX} Security}, 2016.

\bibitem[\protect\citeauthoryear{Weng \bgroup \em et al.\egroup
  }{2018a}]{weng2018towards}
Tsui{-}Wei Weng, Huan Zhang, Hongge Chen, Zhao Song, Cho{-}Jui Hsieh, Luca
  Daniel, Duane~S. Boning, and Inderjit~S. Dhillon.
\newblock Towards fast computation of certified robustness for relu networks.
\newblock In {\em Proceedings of the 35th International Conference on Machine
  Learning, {ICML}}, 2018.

\bibitem[\protect\citeauthoryear{Weng \bgroup \em et al.\egroup
  }{2018b}]{weng2018evaluating}
Tsui{-}Wei Weng, Huan Zhang, Pin{-}Yu Chen, Jinfeng Yi, Dong Su, Yupeng Gao,
  Cho{-}Jui Hsieh, and Luca Daniel.
\newblock Evaluating the robustness of neural networks: An extreme value theory
  approach.
\newblock {\em CoRR}, abs/1801.10578, 2018.

\bibitem[\protect\citeauthoryear{Wicker \bgroup \em et al.\egroup
  }{2018}]{WickerHK18}
Matthew Wicker, Xiaowei Huang, and Marta Kwiatkowska.
\newblock Feature-guided black-box safety testing of deep neural networks.
\newblock In {\em Tools and Algorithms for the Construction and Analysis of
  Systems - 24th International Conference, {TACAS}}, 2018.

\bibitem[\protect\citeauthoryear{Wong and Kolter}{2018}]{WongK18}
Eric Wong and J.~Zico Kolter.
\newblock Provable defenses against adversarial examples via the convex outer
  adversarial polytope.
\newblock In {\em Proceedings of the 35th International Conference on Machine
  Learning, {ICML}}, 2018.

\bibitem[\protect\citeauthoryear{Xu \bgroup \em et al.\egroup }{2016}]{XuQE16}
Weilin Xu, Yanjun Qi, and David Evans.
\newblock Automatically evading classifiers: {A} case study on {PDF} malware
  classifiers.
\newblock In {\em 23rd Annual Network and Distributed System Security
  Symposium, {NDSS}}, 2016.

\bibitem[\protect\citeauthoryear{Xu \bgroup \em et al.\egroup }{2018}]{Xu0Q18}
Weilin Xu, David Evans, and Yanjun Qi.
\newblock Feature squeezing: Detecting adversarial examples in deep neural
  networks.
\newblock In {\em 25th Annual Network and Distributed System Security
  Symposium, {NDSS}}, 2018.

\bibitem[\protect\citeauthoryear{Zantedeschi \bgroup \em et al.\egroup
  }{2017}]{ZantedeschiNR17}
Valentina Zantedeschi, Maria{-}Irina Nicolae, and Ambrish Rawat.
\newblock Efficient defenses against adversarial attacks.
\newblock {\em CoRR}, abs/1707.06728, 2017.

\end{thebibliography}
\bibliographystyle{named}

\end{document}